\documentclass{article}
 
\PassOptionsToPackage{numbers}{natbib}
\usepackage[preprint]{neurips_2026}
 
\usepackage[utf8]{inputenc}
\usepackage[T1]{fontenc}
\usepackage{hyperref}
\usepackage{url}
\usepackage{booktabs}
\usepackage{amsfonts}
\usepackage{nicefrac}
\usepackage{microtype}
\usepackage{xcolor}
\usepackage{amsmath}
\usepackage{makecell}
\usepackage{graphicx}
 
\title{Track2View: 4D-Consistent Camera-Controlled Video Generation via Paired 3D Point Tracks}
 
\author{%
  Feng Qiao\textsuperscript{1} \quad
  Zhaochong An\textsuperscript{2} \quad
  Zhexiao Xiong\textsuperscript{1} \quad
  Serge Belongie\textsuperscript{2} \quad
  Nathan Jacobs\textsuperscript{1} \\[0.5em]
  \textsuperscript{1}Washington University in St.~Louis \quad
  \textsuperscript{2}University of Copenhagen \\
}
 
\begin{document}
 
\maketitle
 
\begin{figure}[h!]
  \centering
  \includegraphics[width=\linewidth]{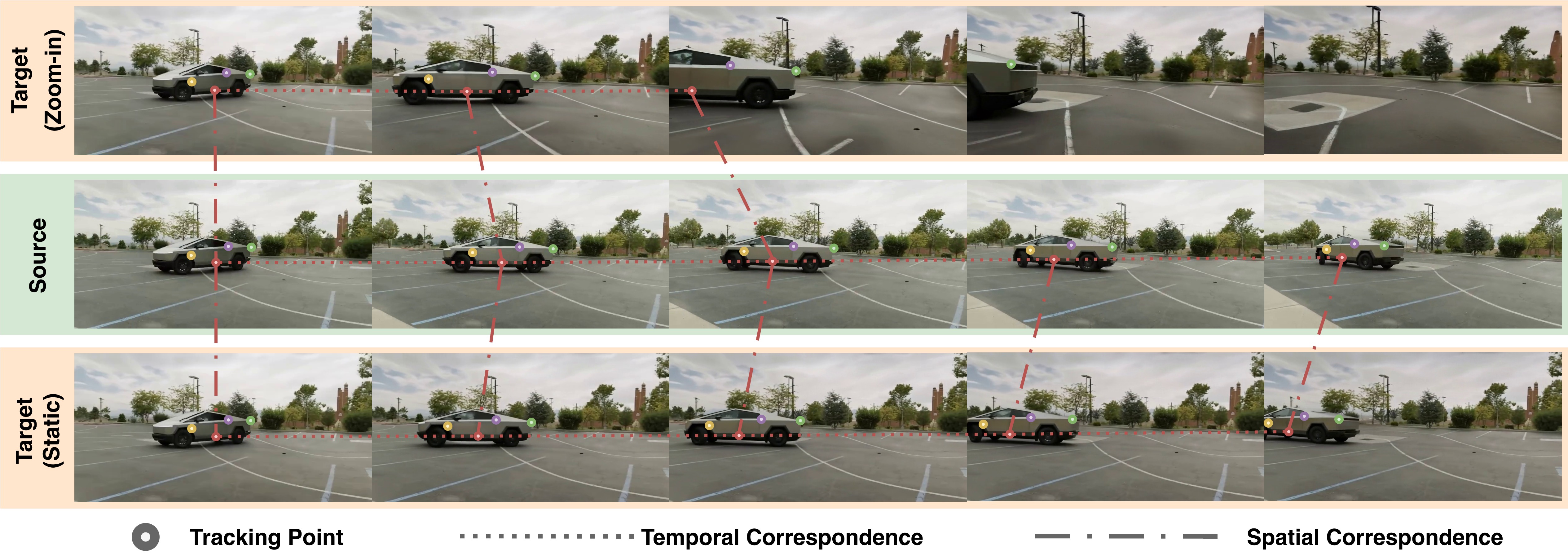}
  \vspace{-6pt}
    \caption{%
      \textbf{Track2View} re-renders a single source video under
      arbitrary target camera trajectories by conditioning on sparse
      3D point tracks.
      %
      %
      Colored markers denote tracked object points; scene-point tracks are also used for conditioning
but omitted from the visualization for clarity. \emph{Spatial correspondence} lines link the same point across views at a fixed
timestep, while \emph{temporal correspondence} lines link it across frames
within a view.
      Together, these correspondences expose the 4D consistency that
      Track2View enforces between source and generated views.
      %
    }
  \label{fig:teaser}
\end{figure}
 
\begin{abstract}
Re-rendering an existing video from a novel camera viewpoint
requires the output to follow the prescribed camera trajectory
while preserving the appearance and dynamics of the original
scene across every frame.
Existing methods rely on per-frame pose embeddings, noisy
point-cloud renderings, or implicit learned correspondences,
none of which provides an explicit, temporally continuous link
between source and target pixels.
We propose Track2View, which conditions a video diffusion
transformer on paired 3D point tracks: sparse trajectories of
scene points projected into both the source and target camera
views.
These tracks provide explicit spatiotemporal correspondences
that are temporally continuous by construction, encoding what
content should appear where and when.
At the core of Track2View is a dual-view track conditioner that
transfers visual context from source to target view through
parameter-free geometric operations and learned temporal
aggregation, ensuring generalization to arbitrary camera
trajectories without memorizing specific motions.
We further introduce a data curation pipeline that extracts
one-to-one track correspondences by running a 3D point tracker
on temporally concatenated multi-camera view pairs.
On a 400-video benchmark spanning static and dynamic scenes,
Track2View achieves state-of-the-art results across visual
quality, view synchronization, and camera accuracy, reducing
rotation error by 30--65\% and translation error by 61--72\%
relative to leading baselines. Project page is available at \href{https://qjizhi.github.io/track2view}{this https URL}.
\end{abstract}

\section{Introduction}

Controlling camera viewpoints in video generation is essential for
applications ranging from filmmaking and virtual reality to robotics
simulation.
Recent advances in video diffusion models have enabled high-quality
text-to-video and image-to-video
synthesis~\cite{blattmann2023stable, wan2025wan, cogvideox,
agarwal2025cosmos}, yet precisely steering the camera trajectory of
a generated video remains a fundamental challenge.
The difficulty is amplified in the video-to-video (V2V) setting,
where the goal is to re-render an existing video from a novel
viewpoint: the generated output must not only follow the prescribed
camera path but also preserve the appearance and dynamics of the
original scene across every frame.
We argue that this task fundamentally demands 4D consistency:
the generated video must be coherent across both space and time
simultaneously.
Yet existing methods fall short of this requirement due to their
choice of camera representation.

Existing camera-controlled V2V methods adopt one of three
strategies, each with notable limitations.
First, Trajectory Attention~\cite{xiao2025trajectory} and
TrajectoryCrafter~\cite{yu2025trajectorycrafter} parameterize
camera motion relative to the source camera per frame, implicitly
assuming a static source camera; when the source camera moves,
they must fall back to image-to-video (I2V) mode, discarding the
source video's temporal context. Their per-frame camera
transformations are also computed independently, with no mechanism
to enforce temporal consistency across frames.
Second, Gen3C~\cite{ren2025gen3c} lifts the source video into a 3D
point cloud via monocular depth estimation and renders each target
frame independently; however, inherently noisy depth estimates
propagate into the rendered guidance, and the per-frame rendering
discards temporal correlations, causing flickering in occluded
regions.
Third, ReCamMaster~\cite{bai2025recammaster} takes a simpler route,
concatenating source and target tokens along the frame dimension
and injecting a learned camera embedding, but delegates the entire
burden of 4D consistency to the model's attention---no explicit
geometric signal links source pixels to their target locations,
and camera accuracy degrades under large viewpoint changes.
In short, none of these methods provides a conditioning signal that
is both geometrically explicit and temporally continuous.

In this work we propose Track2View, a track-conditioned video
re-generation framework that fills this gap.
Our key insight is that 3D point tracks---sparse trajectories of
scene points projected into 2D screen space under both the source
and target camera sequences---naturally encode spatiotemporal
correspondences that are, by construction, temporally continuous.
Tracks are defined in a global coordinate system and remain valid
regardless of the source camera's motion, unlike per-frame relative
parameterizations.
They are also temporally continuous by construction, avoiding the
flickering caused by independent per-frame rendering.
Unlike learned camera embeddings, tracks provide explicit
motion vectors that leave no ambiguity about where each source pixel should appear in the target video.
As illustrated in Fig.~\ref{fig:teaser}, given a source video,
Track2View extracts paired 3D point tracks and uses them to
condition a video diffusion transformer, re-rendering the scene
under diverse target camera trajectories while preserving both
spatial and temporal correspondences with the source.

Our main contributions are as follows:
\begin{itemize}
  \item We introduce Track2View, a V2V framework that conditions on
    paired 3D point tracks for camera control. The tracks encode
    both camera-induced and object-induced displacement, providing
    explicit 4D correspondences between source and target views.
  \item We design a dual-view track conditioner whose geometric
    operations (bilinear sampling and scattering) are entirely
    parameter-free, ensuring generalization to arbitrary camera
    trajectories, while learned temporal aggregation captures
    cross-frame context.
  \item We propose a data curation pipeline that extracts one-to-one
    track correspondences from multi-camera synchronized videos by
    tracking through temporally concatenated view pairs.
  \item On a 400-video benchmark spanning static and dynamic
    scenes, Track2View achieves state-of-the-art results across all
    three evaluation axes, reducing camera rotation error by
    30--65\% and translation error by 61--72\% relative to the best
    existing methods.
\end{itemize}

\section{Related Work}

\subsection{Video Diffusion Models}

Building on denoising diffusion
frameworks~\cite{ho2020denoising,song2020denoising}, early video
generation methods~\cite{ho2022video,animatediff,wang2023lavie}
extend image diffusion backbones with temporal modules. Subsequent
works move to latent video diffusion~\cite{blattmann2023stable},
enabling more efficient spatiotemporal synthesis at higher
resolution and longer durations.
More recently, Diffusion Transformers
(DiT)~\cite{peebles2023scalable} have emerged as a dominant
paradigm, offering unified modeling of spatial and temporal
dependencies through attention-based
architectures~\cite{cogvideox,brooks2024video,kling,wan2025wan,
polyak2024movie,an2025onestory,an2026vggrpo}. These models scale
to large datasets and demonstrate strong generation fidelity and
long video generation ability~\cite{an2026video}.
Despite these advances, existing foundation models primarily focus
on text- or image-conditioned video generation.
However, many emerging applications such as filmmaking and robotics
simulation require explicit control over camera trajectories, as
well as video-to-video re-rendering from novel viewpoints while
preserving scene consistency. This has motivated a growing line of
work toward camera-controllable and geometry-aware video generation.

\subsection{Camera-Controlled and Novel-View Video Generation}

Recent work has increasingly focused on camera-controlled and
novel-view video-to-video generation. Early systems such as
GCD~\cite{gcd} and ReCapture~\cite{recapture} establish the
monocular reshooting setting from a single input video.
Existing approaches can be broadly grouped into three categories,
each with limitations in achieving full spatiotemporal consistency.
Camera-only conditioning approaches inject camera poses or motion
descriptors into a pretrained video
generator~\cite{motionctrl,he2025cameractrl,bahmani2025ac3d,
houtraining,cheong2024boosting,luo2025camclonemaster,ge2026campilot}.
These methods provide only global control signals, leaving dense
source-to-target correspondences implicit.
Some methods~\cite{xiao2025trajectory,yu2025trajectorycrafter}
further parameterize camera motion relative to the source camera
per frame, implicitly assuming a static source camera and
discarding temporal context when the source camera moves.
Rendered-geometry conditioning approaches reconstruct a scene proxy
and render target-view scaffolds to guide
generation~\cite{recapture,ren2025gen3c,
chen2026beyond,cao2026freeorbit4d,qiao2025genstereo}.
While this provides stronger spatial grounding, monocular depth
estimates are inherently noisy, and per-frame rendering discards
temporal correlations, causing flickering and geometric distortions
in occluded regions.
Latent-scene transfer approaches infer view transfer from
source-video tokens or learned scene
latents~\cite{bai2025recammaster,jeong2025reangle,xie2026lavr,
paliwal2026reshoot}.
These methods achieve high visual quality, but the geometric link
between source and target remains implicit---the entire burden of
spatial consistency is delegated to the model's attention, and
camera accuracy degrades under large viewpoint changes.
In contrast, Track2View provides explicit spatiotemporal
correspondence between source and target views through sparse paired 3D
point tracks, which are temporally continuous by construction.
This avoids the noise of rendered geometry, the limitations of
per-frame relative parameterizations, and the implicit nature of
learned embeddings, while maintaining cross-frame consistency
through a shared track representation.

\subsection{Point Tracking and Trajectory Conditioning}

Recent progress in point tracking and feed-forward 4D reconstruction
has made track-based conditioning substantially more practical.
CoTracker~\cite{cotracker} and
CoTracker3~\cite{karaev2025cotracker3} jointly track large sets of
points, improving robustness under occlusion.
SpatialTracker~\cite{spatialtracker} and
SpatialTrackerV2~\cite{xiao2025spatialtrackerv2} further lift
tracking to 3D by jointly modeling scene geometry, camera
ego-motion, and object motion.
In parallel, feed-forward reconstruction
methods~\cite{dust3r_cvpr24,monst3r} recover dense point maps
without per-scene optimization, while more recent
systems~\cite{wang2025vggt,feng2025st4rtrack} unify camera
estimation, point maps, reconstruction, and tracking within a
single feed-forward pipeline.
Building on these advances, recent
work~\cite{dragnuwa,zhang2025tora,xiao2025trajectory,
geng2024motionprompting,fu20243dtrajmaster} has shown that
trajectories can serve as powerful control signals for video
generation, conditioning on 2D or 3D motion paths.
Most recently, Edit-by-Track~\cite{lee2025editbytrack} demonstrated that conditioning on 3D point tracks enables joint camera and object motion editing in V2V generation. Inspired by this, we adapt the track-conditioning paradigm to camera-controlled re-rendering with two key designs.  First, our data curation pipeline (Sec.~\ref{sec:data_curation}) produces naturally synchronized paired tracks, where source track $n$ and target track $n$ correspond to the same 3D point by construction. This one-to-one correspondence enables deterministic bilinear sampling and scattering with no learned parameters in the geometry path, in contrast to the learned cross-attentional sampling/splatting required when tracks are estimated independently from edited 3D point clouds. Second, we adopt SpatialTrackerV2~\cite{xiao2025spatialtrackerv2} with multi-frame query initialization for denser track coverage; Tab.~\ref{tab:abl_query} shows that this improves camera accuracy over single-frame initialization.

\section{Method}
\label{sec:method}
The camera-controlled video re-rendering task can be defined
as: given a source video
$V_\text{src} \in \mathbb{R}^{F \times H \times W \times 3}$ and a
user-specified target camera trajectory
$\mathcal{P}_\text{tgt}$, the goal is to generate a
frame-synchronized target video
$V_\text{tgt} \in \mathbb{R}^{F \times H \times W \times 3}$ that
faithfully reproduces the scene content and dynamics of
$V_\text{src}$ as observed from the target viewpoint, where each
target frame $V_\text{tgt}^t$ corresponds to the same time step as
$V_\text{src}^t$.
To enable track-based conditioning, we use
SpatialTrackerV2~\cite{xiao2025spatialtrackerv2} to jointly
estimate camera poses $\mathcal{P}_\text{src}$ and 3D point tracks
$\mathcal{T}_\text{src} \in \mathbb{R}^{F \times N \times 3}$
from the source video, and reproject
$\mathcal{T}_\text{src}$ under $\mathcal{P}_\text{tgt}$ to obtain
the corresponding target tracks $\mathcal{T}_\text{tgt}$.

\subsection{Framework}
\label{sec:framework}

\begin{figure}[t]
  \centering
  \includegraphics[width=\linewidth]{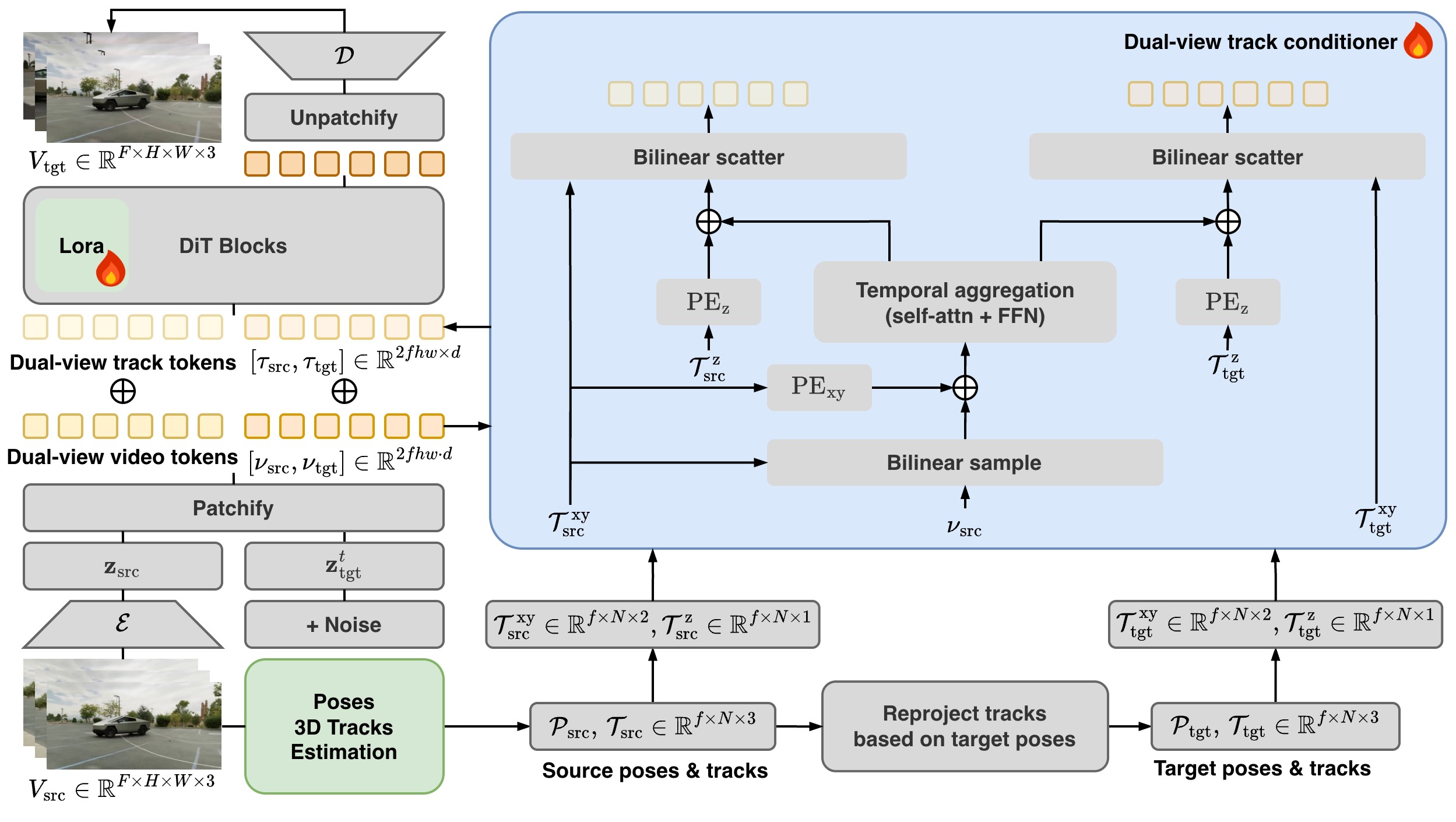}
  \caption{%
    \textbf{Overview of Track2View.}
    $\oplus$ denotes element-wise addition;
    $\mathrm{PE}$ denotes Fourier positional encoding;
    $N$ denotes the number of point tracks.
  }
  \label{fig:framework}
\end{figure}

As illustrated in Fig.~\ref{fig:framework}, the estimated tracks
are projected into paired screen-space coordinates
$(\mathcal{T}_\text{src}^{xy},
  \mathcal{T}_\text{tgt}^{xy}) \in \mathbb{R}^{f \times N \times 2}$
and per-view depths
$(\mathcal{T}_\text{src}^{z},
  \mathcal{T}_\text{tgt}^{z}) \in \mathbb{R}^{f \times N \times 1}$.

The source video is encoded by a 3D variational autoencoder (VAE) encoder $\mathcal{E}$ into
a latent $\mathbf{z}_\text{src} = \mathcal{E}(V_\text{src})$.
The 3D VAE temporally compresses the video by a factor of
$4{\times}$, yielding $f{=}21$ latent frames from $F{=}81$ input
frames; point tracks are subsampled at the same stride to align
with the latent frame rate.
The latent is patchified into source tokens $\nu_\text{src}$.
A noisy target latent $\mathbf{z}_\text{tgt}^t$ is sampled by
adding noise at diffusion timestep $t$, and patchified into target
tokens $\nu_\text{tgt}$.
The two are concatenated along the frame dimension to form
dual-view video tokens
$[\nu_\text{src}, \nu_\text{tgt}] \in
\mathbb{R}^{2fhw \times d}$.
Our dual-view track conditioner
(Sec.~\ref{sec:track_conditioner}) then produces track tokens
$[\tau_\text{src}, \tau_\text{tgt}] \in
\mathbb{R}^{2fhw \times d}$, which are added element-wise to the
video tokens before they are processed by the DiT blocks.
We fine-tune the pretrained DiT with Low-Rank Adaptation (LoRA) adapters while keeping
all other parameters frozen, allowing the model to learn
track-conditioned generation without catastrophic forgetting of
the base video prior.
The DiT denoises the target portion of the concatenated tokens,
producing a clean latent $\mathbf{z}_\text{tgt}^0$ that is
unpatchified and decoded by the 3D VAE decoder $\mathcal{D}$ to
yield the final output video
$V_\text{tgt} = \mathcal{D}(\mathbf{z}_\text{tgt}^0)$.

\subsection{Dual-View Track Conditioner}
\label{sec:track_conditioner}

The dual-view track conditioner converts sparse paired 3D tracks
into dense conditioning tokens that inject geometric correspondence
into the diffusion process.
All geometric operations---sampling and scattering---use
deterministic bilinear interpolation with no learnable parameters,
ensuring that camera geometry is encoded directly rather than
memorized; only the temporal aggregation and feature projections
are learned, keeping the module lightweight and generalizable to
unseen trajectories.

Given $N$ 3D scene points tracked across $f$ frames, we project
them into both the source and target camera views using their
respective camera parameters, obtaining normalized 2D coordinates
$\mathcal{T}^{xy} \in \mathbb{R}^{f \times N \times 2}$
and camera-space depths
$\mathcal{T}^{z} \in \mathbb{R}^{f \times N \times 1}$
for each view, along with a validity mask
$\mathbf{m} \in \{0,1\}^{f \times N}$ marking tracks with
positive depth that fall within the image bounds.

We extract per-track visual features by sampling from the source
video tokens $\nu_\text{src} \in \mathbb{R}^{f \times hw \times d}$
at the projected source coordinates
$\mathcal{T}_\text{src}^{xy}$ via bilinear grid sampling,
yielding sparse track features
$\hat{\tau} \in \mathbb{R}^{f \times N \times d}$.
We then add a Fourier positional encoding of the sampling
coordinates and apply a residual MLP for refinement:
\begin{equation}
  \hat{\tau} \leftarrow
  \hat{\tau}
  + \mathrm{PE}_{xy}(\mathcal{T}_\text{src}^{xy})
  + \mathrm{MLP}(\hat{\tau}
    + \mathrm{PE}_{xy}(\mathcal{T}_\text{src}^{xy})).
\end{equation}
Invalid tracks (outside the source view) are zeroed out.

Each 3D point is observed across multiple frames, providing
complementary visual information from different time steps.
We aggregate this per track across the temporal dimension using
an $L$-layer transformer encoder ($L{=}8$ by default) with a
padding mask that excludes frames where a track is occluded or
out of view:
\begin{equation}
  \bar{\tau} = \mathrm{TemporalAgg}(\hat{\tau}),
  \quad \bar{\tau} \in \mathbb{R}^{f \times N \times d}.
\end{equation}
This propagates visual context from frames where a point is
visible to frames where it is not, establishing temporal
correspondences without relying on the DiT's own attention.

After temporal aggregation, we inject view-specific 3D structure by encoding the
inverse depth $1/\mathcal{T}^z$ (disparity), jointly normalized across the source
and target views to preserve their relative scale, via Fourier features:
\begin{equation}
  v_\text{src} = \bar{\tau}
    + \mathrm{PE}_z(\mathcal{T}_\text{src}^{z}),
  \quad
  v_\text{tgt} = \bar{\tau}
    + \mathrm{PE}_z(\mathcal{T}_\text{tgt}^{z}).
\end{equation}
The temporal features $\bar{\tau}$ are shared between both views
since they describe the same 3D points, while the depth encodings
differentiate the two viewpoints, allowing the model to reason
about parallax and occlusion.

Finally, we scatter the per-track features back to dense spatial
grids using the inverse of bilinear interpolation, where each track
distributes its feature to the four nearest grid cells with
bilinear weights, and cells receiving multiple contributions are
averaged:
\begin{equation}
  \tau_\text{src} = \mathrm{Scatter}(v_\text{src},\;
    \mathcal{T}_\text{src}^{xy}),
  \quad
  \tau_\text{tgt} = \mathrm{Scatter}(v_\text{tgt},\;
    \mathcal{T}_\text{tgt}^{xy}).
\end{equation}
The resulting dual-view track tokens
$[\tau_\text{src}, \tau_\text{tgt}]
\in \mathbb{R}^{2fhw \times d}$ are added element-wise to the
video tokens before the DiT blocks.
Because both views share the same underlying 3D points, this
mechanism naturally enforces consistent rendering across
viewpoints.

During training, we apply source frame dropout (randomly zeroing
up to 50\% of source frames) and track subsampling (retaining
50--100\% of tracks) to improve robustness to varying input
conditions.

\subsection{Training Data Curation}
\label{sec:data_curation}

Our training requires paired videos of the same dynamic scene
captured from two different camera viewpoints, together with
one-to-one 3D point track correspondences between them.
We source our data from the MultiCamVideo dataset of
ReCamMaster~\cite{bai2025recammaster}, which consists of
synthetic scenes rendered in Unreal Engine 5 where all
virtual cameras are initialized at the same position,
ensuring that they share an identical first frame.
For each scene, we randomly select two cameras $A$ and $B$ to
form a training pair, treating camera $A$ as the source and
camera $B$ as the target, where each video contains $F{=}81$
frames.

\begin{figure}[t]
  \centering
  \includegraphics[width=\linewidth]{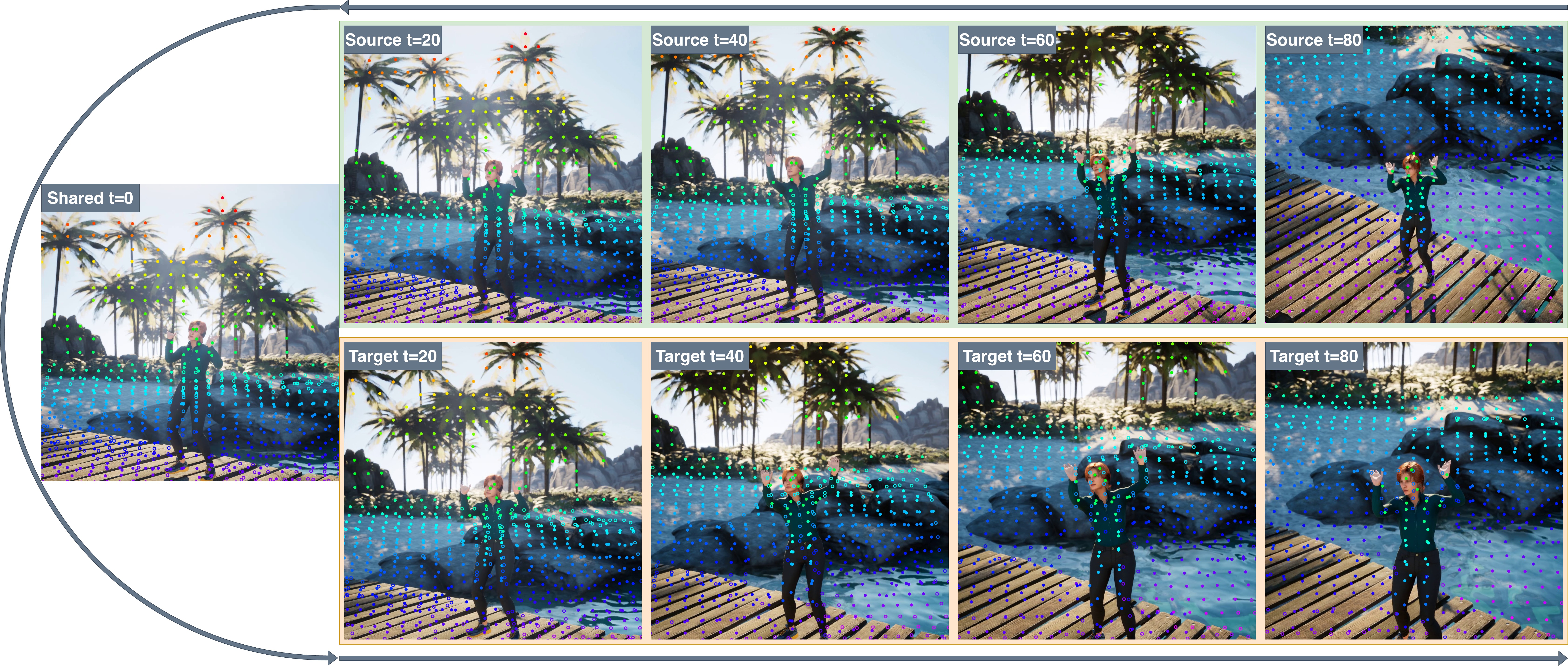}
  \caption{%
    \textbf{Paired track extraction.}
    We temporally reverse the source video (camera $A$) and
    concatenate it with the target video (camera $B$), forming a
    161-frame sequence with a shared first frame.
    Top: source frames (reversed back to original order).
    Bottom: target frames.
    Colored dots denote 3D points tracked by
    SpatialTrackerV2~\cite{xiao2025spatialtrackerv2} with queries
    at frames $0$ and $80$; matching colors across rows indicate
    one-to-one correspondences.
  }
  \label{fig:data_curation}
\end{figure}

To obtain one-to-one point track correspondences between the
source and target videos, we temporally reverse camera $A$ and
concatenate it with camera $B$, yielding a single
$2F{-}1{=}161$-frame sequence.
Because all cameras share the same first frame, the last frame
of the reversed camera $A$ (i.e., frame 0 of the original) is
identical to the first frame of camera $B$, producing a seamless
transition at the concatenation boundary.
We run SpatialTrackerV2~\cite{xiao2025spatialtrackerv2} on the
concatenated sequence, initializing query points at frames $0$
and $80$ only---we do not place queries beyond frame $80$
because at inference time only the source video (81 frames) is
available.
Since the tracker follows each point continuously through the
entire sequence, the resulting tracks naturally span both the
source and target segments, establishing one-to-one
correspondences without any explicit matching.
The tracks from the first 81 frames (reversed camera $A$) are
re-reversed to recover the original temporal order, giving us
paired track sets
$(\mathcal{T}_\text{src}, \mathcal{T}_\text{tgt}) \in
\mathbb{R}^{2 \times F \times N \times 3}$
ready for training.
As shown in Fig.~\ref{fig:data_curation}, the extracted tracks
maintain consistent point identities across the source and target
views.
At inference time, the target tracks are obtained by reprojecting
the source 3D point tracks under the user-specified target camera
poses, requiring no additional tracking.
We ablate the effect of query frame selection in
Tab.~\ref{tab:abl_query}.

\subsection{Training Details}
\label{sec:training_details}

We build Track2View on WAN-2.1~\cite{wan2025wan}, a pretrained
text-to-video diffusion transformer that generates 81-frame videos
at $480 \times 832$ resolution.
The dual-view track conditioner is trained from scratch with full
parameter updates, while the pretrained DiT blocks are adapted
using LoRA~\cite{hu2022lora} with rank $r{=}64$
and $\alpha{=}64$, applied to the query, key, value, output
projection, and feed-forward layers of each transformer block.
All other parameters of the base model remain frozen.

By default we set $N{=}1152$, with half of the queries placed at the first frame and half at the middle frame ($t{=}80$) of the temporally concatenated $2F{-}1$ sequence used for paired-track extraction (Sec.~\ref{sec:data_curation}).
During training, the source latent
$\mathbf{z}_\text{src}$ remains clean (no noise added), while noise
is applied only to the target latent $\mathbf{z}_\text{tgt}$.
The training objective is the standard flow-matching
loss~\cite{lipman2023flow} computed on the target portion of the
concatenated sequence only, ensuring the model learns to denoise
the target view conditioned on the clean source.
We train with AdamW at a learning rate of $1 \times 10^{-5}$ in
bfloat16 precision using DeepSpeed ZeRO Stage 1.
Training takes approximately 576 GPU-hours on 4 A100 GPUs with a
batch size of 2 per GPU over 216K optimization steps.

At inference time, we use 50 denoising steps with classifier-free 
guidance scale $5.0$ applied to the text prompt, following the 
default sampling configuration of the WAN-2.1 base model.

\section{Experiments}

\subsection{Dataset and Evaluation Protocol}
\label{sec:eval_protocol}

We evaluate on the RealCam-Vid benchmark~\cite{zheng2025realcamvid}.
Our evaluation set comprises 400 videos: 200 static scenes sourced
from RealEstate10K~\cite{zhou2018stereo} and
200 dynamic scenes from MiraData~\cite{ju2024miradata}, enabling separate assessment under different
scene conditions.
FID reference features are pre-extracted from 33{,}388 real frames
(Inception-v3) sampled from the RealCam-Vid test set. We use the text captions provided by the benchmark as the text conditioning signal.

\subsection{Metrics}

We evaluate along three complementary axes.

Visual quality is measured by
FID$\downarrow$ (Fr\'echet Inception Distance), which computes the
Fr\'echet distance between Inception-v3 features of generated frames
and those of the real video pool;
CLIP-T$\uparrow$, which measures text-video semantic alignment via
CLIP ViT-B/32 between generated frames and the text prompt; and
CLIP-F$\uparrow$, which measures temporal consistency as the average
CLIP cosine similarity between consecutive generated frames.

View synchronization is measured by
CLIP-V$\uparrow$, the average CLIP cosine similarity between source
and generated frames at the same timestamp; and
Mat.Pix.~(K)$\uparrow$, which counts confidently matched keypoints
(in thousands) between source and generated frame pairs using
GIM-LightGlue~\cite{shen2024gim}, providing a geometry-aware
correspondence signal.

Camera accuracy is measured by
RotErr$\downarrow$ and TransErr$\downarrow$, following the protocol
of CameraCtrl~\cite{he2025cameractrl}.
We extract camera poses from generated videos using
GLOMAP~\cite{pan2024glomap} (via COLMAP~\cite{schoenberger2016sfm}
global mapper).
RotErr reports the mean geodesic distance (in degrees) between
ground-truth and estimated rotation matrices, and TransErr reports
the mean $\ell_2$ distance between translation vectors after
scale alignment to the first two frames. Both metrics compare extracted poses against the ground-truth
target trajectory.

We additionally report six VBench~\cite{huang2024vbench} dimensions---Aesthetic Quality, Imaging Quality, Temporal Flickering, Motion
Smoothness, Subject Consistency, and Background Consistency---to assess the perceptual and temporal quality of the generated videos.

\subsection{Baselines}

We compare Track2View against four state-of-the-art camera-controlled
video generation methods:
Trajectory Attention~\cite{xiao2025trajectory},
TrajectoryCrafter~\cite{yu2025trajectorycrafter},
Gen3C~\cite{ren2025gen3c}, and
ReCamMaster~\cite{bai2025recammaster}.
Since Trajectory Attention and TrajectoryCrafter parameterize camera
motion relative to the source frame, their camera accuracy is
evaluated in I2V mode (denoted $^*$ in Tab.~\ref{tab:main}).
For fair comparison, we also evaluate Track2View at 25 and 49 frames
to match their respective output lengths.

\subsection{Quantitative Comparisons}

Quantitative results are reported in Tab.~\ref{tab:main}.
Since Trajectory Attention and TrajectoryCrafter generate 25 and 49 frames respectively, we also evaluate Track2View at matching frame counts by taking the first 25 or 49 frames of our 81-frame output.
Track2View consistently outperforms all baselines across the three evaluation axes.

In visual quality, Track2View achieves the lowest FID at every frame length (e.g., 26.82 vs.\ 30.32 for Gen3C and 33.85 for ReCamMaster at 81 frames), while maintaining comparable or better CLIP-T and CLIP-F scores, confirming that the gains do not come at the cost of semantic fidelity.

For view synchronization, Track2View achieves the highest CLIP-V
(93.22 vs.\ 92.93 for ReCamMaster) and Mat.Pix. (0.695K vs.\
0.644K for Gen3C and 0.579K for ReCamMaster) at 81 frames,
confirming stronger geometric correspondence between source and
generated views.
Gains are most pronounced at 25 frames, where Mat.Pix.\ reaches
1.070K compared to 0.826K for Trajectory Attention (+29.5\%).

Camera accuracy sees the most significant improvements: RotErr drops by 30--65\% and TransErr by 61--72\% relative to each respective baseline across all frame lengths. The largest relative gains appear at 25 frames (RotErr $1.24^\circ$ vs.\ $3.54^\circ$, $-$65\%), while absolute errors naturally grow at 81 frames as pose estimation errors accumulate over longer sequences; nonetheless, Track2View still reduces RotErr by 30\% and TransErr by 61\% compared to the strongest baseline at this length.

We further evaluate on six VBench~\cite{huang2024vbench} dimensions
(Tab.~\ref{tab:vbench}).
Track2View achieves the best or second-best score on all dimensions
across all frame lengths, demonstrating that improved camera
controllability does not come at the cost of perceptual video
quality.

\begin{table}[t]
  \caption{%
    Quantitative comparison on the RealCam-Vid benchmark (400 videos).
    $^*$Camera accuracy evaluated in I2V mode (camera trajectory
    relative to the source frame).
  }
  \label{tab:main}
  \centering
  \scriptsize
  \setlength{\tabcolsep}{4pt}
  \begin{tabular}{l c ccc cc cc}
    \toprule
    & & \multicolumn{3}{c}{\textit{Visual Quality}}
    & \multicolumn{2}{c}{\textit{View Sync.}}
    & \multicolumn{2}{c}{\textit{Cam. Acc.}} \\
    \cmidrule(lr){3-5}\cmidrule(lr){6-7}\cmidrule(lr){8-9}
    Method & Frames
      & FID$\downarrow$ & CLIP-T$\uparrow$ & CLIP-F$\uparrow$
      & CLIP-V$\uparrow$ & Mat.Pix.(K)$\uparrow$
      & RotErr($^\circ$)$\downarrow$ & TransErr$\downarrow$ \\
    \midrule
    Trajectory Attention$^*$~\cite{xiao2025trajectory} & 25
      & 38.25 & 28.57 & 98.49 & 94.87 & 0.826 & 3.54 & 0.309 \\
    Track2View (Ours) & 25
      & \textbf{33.95} & \textbf{29.06} & \textbf{99.28}
      & \textbf{95.81} & \textbf{1.070}
      & \textbf{1.24}{\scriptsize\,(-65\%)}
      & \textbf{0.085}{\scriptsize\,(-72\%)} \\
    \midrule
    TrajectoryCrafter$^*$~\cite{yu2025trajectorycrafter} & 49
      & 29.34 & 28.87 & 98.70 & 94.70 & 0.737 & 2.12 & 0.721 \\
    Track2View (Ours) & 49
      & \textbf{29.30} & \textbf{29.07} & \textbf{99.35}
      & \textbf{94.71} & \textbf{0.876}
      & \textbf{1.31}{\scriptsize\,(-38\%)}
      & \textbf{0.280}{\scriptsize\,(-61\%)} \\
    \midrule
    Gen3C~\cite{ren2025gen3c} & 81
      & 30.32 & 28.55 & 99.04 & 92.47 & 0.644 & 4.21 & 3.715 \\
    ReCamMaster~\cite{bai2025recammaster} & 81
      & 33.85 & 28.48 & 99.33 & 92.93 & 0.579 & 2.20 & 2.096 \\
    Track2View (Ours) & 81
      & \textbf{26.82} & \textbf{28.89} & \textbf{99.38}
      & \textbf{93.22} & \textbf{0.695}
      & \textbf{1.55}{\scriptsize\,(-30\%)}
      & \textbf{0.818}{\scriptsize\,(-61\%)} \\
    \bottomrule
  \end{tabular}
\end{table}

\begin{table}[t]
  \caption{%
    VBench~\cite{huang2024vbench} evaluation.
  }
  \label{tab:vbench}
  \centering
  \scriptsize
  \setlength{\tabcolsep}{4pt}
  \begin{tabular}{l c cccccc}
    \toprule
    Method & Frames
      & \makecell{Aesthetic\\Quality$\uparrow$}
      & \makecell{Imaging\\Quality$\uparrow$}
      & \makecell{Temporal\\Flickering$\uparrow$}
      & \makecell{Motion\\Smoothness$\uparrow$}
      & \makecell{Subject\\Consistency$\uparrow$}
      & \makecell{Background\\Consistency$\uparrow$} \\
    \midrule
    Trajectory Attention~\cite{xiao2025trajectory} & 25
      & 50.37 & 66.00 & 96.00 & 98.78 & 96.22 & 94.82 \\
    Track2View (Ours) & 25
      & \textbf{53.54} & \textbf{70.28} & \textbf{96.94}
      & \textbf{99.26} & \textbf{97.66} & \textbf{95.72} \\
    \midrule
    TrajectoryCrafter~\cite{yu2025trajectorycrafter} & 49
      & 53.44 & 70.01 & 95.28 & 98.63 & 93.95 & 93.89 \\
    Track2View (Ours) & 49
      & \textbf{53.87} & \textbf{70.11} & \textbf{97.02}
      & \textbf{99.28} & \textbf{96.67} & \textbf{94.92} \\
    \midrule
    Gen3C~\cite{ren2025gen3c} & 81
      & 51.53 & \underline{68.83} & 96.49
      & 99.12 & 93.75 & 92.36 \\
    ReCamMaster~\cite{bai2025recammaster} & 81
      & \underline{52.29} & 67.52 & \textbf{97.23}
      & \textbf{99.29} & \underline{95.26} & \underline{93.47} \\
    Track2View (Ours) & 81
      & \textbf{53.68} & \textbf{69.85} & \underline{97.13}
      & \textbf{99.29} & \textbf{95.33} & \textbf{94.22} \\
    \bottomrule
  \end{tabular}
\end{table}

\subsection{Qualitative Comparisons}

\begin{figure*}[t]
  \centering
  \includegraphics[width=\textwidth]{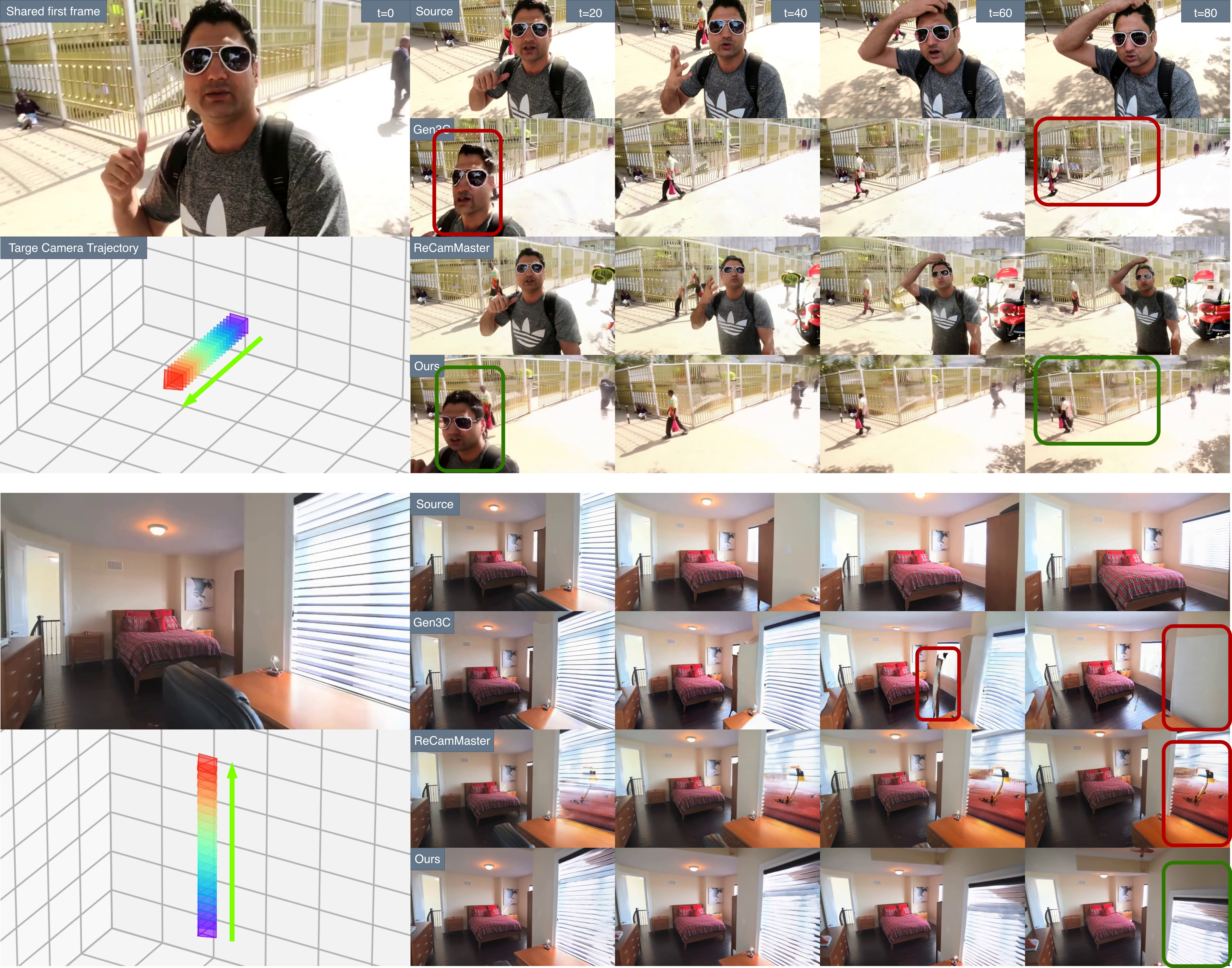}
  \caption{%
    Qualitative comparison on a dynamic outdoor scene (top) and a
    static indoor scene (bottom).
    Red boxes highlight artifacts or hallucinated content; green
    boxes indicate faithful generation by our method.
  }
  \label{fig:qualitative}
\end{figure*}

Fig.~\ref{fig:qualitative} presents qualitative comparisons
with Gen3C~\cite{ren2025gen3c} and
ReCamMaster~\cite{bai2025recammaster} on two representative
scenes.
In the dynamic outdoor scene (top), Gen3C introduces floating
artifacts and inconsistent foreground objects and background geometry due to noisy
per-frame depth warping, while ReCamMaster fails to follow the prescribed camera trajectory and hallucinates objects
absent from the source video (e.g., a red scooter in the
background).
In the static indoor scene (bottom), both baselines exhibit
progressive content drift as the camera moves upward---new
objects appear or disappear in later frames, breaking 4D
consistency.
Our method avoids these failure modes by conditioning on paired
3D tracks that anchor generated content to the source video's
geometry, producing temporally stable outputs that preserve
scene content faithfully even under large camera motions.

\subsection{Ablation Studies}

We ablate two key design choices of Track2View: the query strategy
for initializing point tracks (Tab.~\ref{tab:abl_query}) and
the depth of the temporal aggregation module
(Tab.~\ref{tab:abl_temporal}).
All ablations are evaluated on the full 400-video benchmark at 81
frames.

\begin{table}[t]
  \begin{minipage}[t]{0.48\linewidth}
    \centering
    \caption{Ablation on query strategy.}
    \label{tab:abl_query}
    \scriptsize
    \setlength{\tabcolsep}{3pt}
    \begin{tabular}{lcccc}
      \toprule
      Configuration
        & CLIP-V$\uparrow$
        & Mat.Pix.$\uparrow$
        & RotErr$\downarrow$
        & TransErr$\downarrow$ \\
      \midrule
      First only (576)
        & 92.79 & 0.682 & 2.36 & \textbf{1.574} \\
      Rand.\ $\nicefrac{1}{2}$ (576)
        & 92.79 & 0.681 & \textbf{2.24} & 1.797 \\
      Rand.\ $\nicefrac{1}{4}$ (288)
        & \textbf{92.80} & 0.680 & 2.36 & 2.440 \\
      All (1152)
        & \textbf{92.80} & \textbf{0.685} & \textbf{2.24}
        & \underline{1.681} \\
      \bottomrule
    \end{tabular}
  \end{minipage}
  \hfill
  \begin{minipage}[t]{0.48\linewidth}
    \centering
    \caption{Ablation on temporal aggregation depth.}
    \label{tab:abl_temporal}
    \scriptsize
    \setlength{\tabcolsep}{3pt}
    \begin{tabular}{lcccc}
      \toprule
      Configuration
        & CLIP-V$\uparrow$
        & Mat.Pix.$\uparrow$
        & RotErr$\downarrow$
        & TransErr$\downarrow$ \\
      \midrule
      2 layers
        & 92.80 & 0.685 & 2.24 & 1.681 \\
      4 layers
        & 92.52 & 0.679 & 1.97 & 1.214 \\
      8 layers
        & \textbf{93.22} & \textbf{0.695} & \textbf{1.55}
        & \textbf{0.818} \\
      \bottomrule
    \end{tabular}
  \end{minipage}
\end{table}

Tab.~\ref{tab:abl_query} compares different strategies for
selecting query frames when initializing tracks with
SpatialTrackerV2~\cite{xiao2025spatialtrackerv2}.
The ``All'' configuration (queries
at both the first and middle frames, $1152$ tracks) yields the best Mat.Pix.\ ($0.685$) and
tied-best RotErr ($2.24^\circ$). While ``First only'' achieves marginally lower TransErr
($1.574$ vs.\ $1.681$) under the shallow 2-layer aggregation used in this ablation, the
mid-sequence queries in ``All'' cover scene content that becomes visible only after the first
frame---e.g., regions revealed by camera motion or de-occlusion---which we find essential once
combined with the deeper 8-layer aggregation in our final model. Aggressively subsampling to
$288$ tracks (``Rand.\ 1/4'') degrades TransErr to $2.440$, indicating that sufficient track
density is also important for precise camera control. We adopt ``All'' as our default. All
configurations in this table use 2 temporal aggregation layers.

Tab.~\ref{tab:abl_temporal} varies the number of self-attention layers in the temporal
aggregation module. Camera accuracy improves monotonically with depth: RotErr drops from
$2.24^\circ$ to $1.55^\circ$ ($-31\%$) and TransErr from $1.681$ to $0.818$ ($-51\%$) as we go
from 2 to 8 layers. Overall, deeper temporal aggregation more effectively
propagates visual context across frames, and we use 8 layers as our default.

\section{Conclusion}

We presented Track2View, a camera-controlled video re-generation
framework that conditions a video diffusion transformer on paired
3D point tracks.
By representing camera motion as sparse trajectories projected
into both source and target screen spaces, our method provides
explicit, temporally continuous correspondences that existing
pose-based and rendering-based approaches lack.
The dual-view track conditioner uses parameter-free bilinear
sampling and scattering for geometry encoding, combined with
learned temporal aggregation for cross-frame context propagation.
Experiments on a 400-video benchmark spanning static and dynamic
scenes demonstrate that Track2View achieves state-of-the-art
results across visual quality, view synchronization, and camera
accuracy, reducing rotation and translation errors by 30--65\%
and 61--72\% respectively compared to leading baselines.

\paragraph{Limitations.}
Track2View has two main limitations.
First, our method relies on the quality of the upstream 3D point 
tracker: scenes where tracking fails may degrade conditioning 
and generation quality.
Second, although our model is trained entirely on synthetic 
Unreal-Engine-5 data and transfers well to real-world videos 
(RealEstate10K, MiraData), its behavior under extreme 
out-of-distribution conditions remains uncharacterized.
Improving robustness to imperfect tracks, for instance through 
confidence-weighted conditioning, is a promising direction for 
future work.

\bibliographystyle{plainnat}
\bibliography{references}


\end{document}